\documentclass[letterpaper, 10 pt, conference]{ieeeconf}  % Comment this line out if you need a4paper

\IEEEoverridecommandlockouts                              % This command is only needed if 
\usepackage{graphics} % for pdf, bitmapped graphics files
\usepackage{epsfig} % for postscript graphics files
\usepackage{times} % assumes new font selection scheme installed
\usepackage{amsmath} % assumes amsmath package installed
\usepackage{amssymb}  % assumes amsmath package installed
\usepackage{balance}
\usepackage[dvipsnames,table,xcdraw]{xcolor}
\usepackage{hyperref}
\hypersetup{
    colorlinks,
    linkcolor={red!45!black},
    citecolor={green!35!black},
    urlcolor={blue!80!black}
}
\usepackage[font=small,labelfont=bf]{caption, subcaption}
\usepackage{graphicx}

\usepackage{multirow}
\usepackage[table,xcdraw]{xcolor}

\usepackage[ruled]{algorithm2e}

\SetCommentSty{mycommfont}

\usepackage{todonotes}
\usepackage{balance}

\title{\LARGE \bf
Learning on the Job: Long-Term Behavioural Adaptation in Human-Robot Interactions
}

\author{Francesco Del Duchetto and Marc Hanheide% <-this % stops a space
\thanks{This work was supported by the Lincolnshire County Council.}% <-this % stops a space
\thanks{Authors are with Lincoln Center for Autonomous Systems (L-CAS), School of Computer Science, University of Lincoln, United Kingdom. Emails: {\tt \footnotesize \{fdelduchetto, mhanheide\}@lincoln.ac.uk} }%
}

\begin{document}
\maketitle
\thispagestyle{empty}
\pagestyle{empty}
%%%%%%%%%%%%%%%%%%%%%%%%%%%%%%%%%%%%%%%%%%%%%%%%%%%%%%%%%%%%%%%%%%%%%%%%%%%%%%%%
\begin{abstract}
% Successfully deploying service robots in public spaces require a wide range of capabilities that are still out of reach for today's state-of-the-art. One such ability is the interaction with users.
In this work, we propose a framework for allowing autonomous robots deployed for extended periods of time in public spaces to adapt their own behaviour online from user interactions.
The robot behaviour planning is embedded in a Reinforcement Learning (RL) framework, where the objective is maximising the level of overall user engagement during the interactions.
We use the Upper-Confidence-Bound Value-Iteration (UCBVI) algorithm, which gives a helpful way of managing the exploration-exploitation trade-off for real-time interactions. 
An engagement model trained end-to-end generates the reward function in real-time during policy execution.
We test this approach in a public museum in Lincoln (UK), where the robot is deployed as a tour guide for the visitors. 
Results show that after a couple of months of exploration, the robot policy learned to maintain the engagement of users for longer, with an increase of 22.8\% over the initial static policy in the number of items visited during the tour and a 30\% increase in the probability of completing the tour. 
This work is a promising step toward behavioural adaptation in long-term scenarios for robotics applications in social settings.
\end{abstract}

%%%%%%%%%%%%%%%%%%%%%%%%%%%%%%%%%%%%%%%%%%%%%%%%%%%%%%%%%%%%%%%%%%%%%%%%%%%%%%%%
\section{INTRODUCTION}
\label{sec:intro}

The ability to maintain user engagement during an interaction is an essential ability for a robot designed to be deployed in a social scenario. 
In order to do so, a robot should be able to assess the users' state and to learn from experience how the actions it takes affect that same users' state and the ongoing interaction.
Reinforcement Learning (RL) techniques are a special case of Machine Learning algorithms that deals exactly with those scenarios where the ``goodness'' of the actions an agent can take is not known in advance, and exploration is required.
The goal is to find the best sequence of actions to maximise a certain objective, which is manifested through rewards.
However difficult this may seem for scenarios where the goal is well defined, it becomes even more challenging for social scenarios where the objective is expressed by the users' internal state and can only be estimated from sensors or proxy variables (like the duration of the interaction).

\begin{figure}[t]
\centering
    \includegraphics[width=\columnwidth]{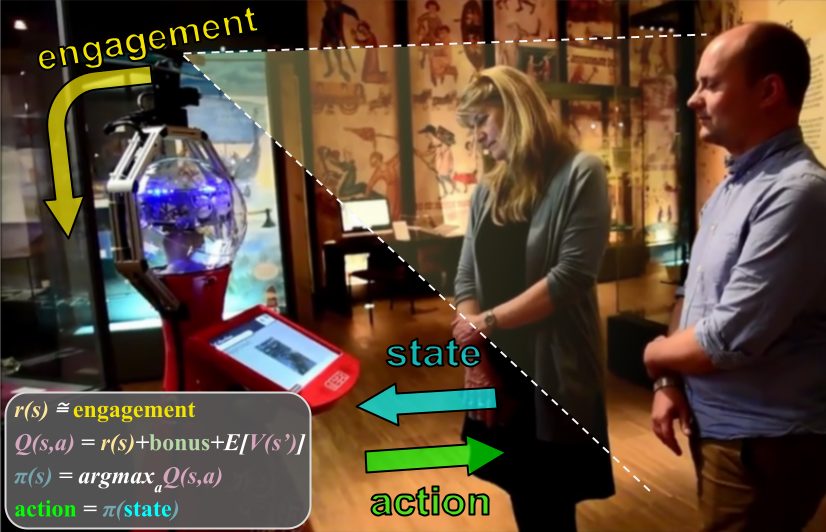}
    % \caption{Caption}
    % \label{fig:my_label}
    \caption{A picture of Lindsey in the archaeological gallery of The Collection museum during a tour guide interaction. The robot learning framework detects user engagement online and optimises its interaction policy for maximising such overall engagement using Reinforcement Learning. }
    \label{fig:arc_gallery}
\end{figure}

In this paper, we approach this problem by enabling our robot Lindsey to estimate the users' engagement during the interaction and by allowing it to learn, through RL, the actions to maximise such engagement.
We build on our previous work \cite{del2019lindsey} where we described and analysed the long-term deployment, which is still ongoing with the present work, of our robot in a public museum where it serves as a tour guide to the visitors. 
For detecting the users' engagement we use our regression model, proposed in \cite{del2020you}, which provides a single scalar engagement from standard video streams
obtained from the point of view of the interacting robot.
Figure \ref{fig:arc_gallery} shows Lindsey the robot in the museum while interacting with the users and depicts the concept for our proposed learning framework.  
 
By attempting to learn in a long-term scenario using only the users' engagement as guidance, we provide a proof-of-concept for a framework to enable behavioural adaptation in social robots. 
Given that the learning is guided by the users' state, manifested by their expressed engagement, this will allow the robot to adapt to the users' preferences, which cannot be known or programmed in advance.
Finally, with our proposed framework we set out to test the hypotheses of whether \emph{ the engagement value estimated during the interactions a sufficient feedback signal for driving the in-situ learning of the robot social behaviours}; and if \emph{it leads to a more sustained interaction with the users}.

\section{RELATED WORK}
% \todo[inline]{this section is copied from our prev papers so it may need to be checked and updated.}

\subsection{Long-Term Deployment in Public Spaces}

Previous work have addressed the issues of deploying robots in public environments for long periods of time taking into account the necessity of interacting with people.
% In the past, several works have addressed the problem of Long-Term Autonomy (LTA) in robotics. 
Past long-term deployments in public spaces \cite{meeussen2011long, biswas20161} have identified the interaction with humans a necessity for recovering for failures and performing tasks that the robot was not able to.
% Meeussen et al. \cite{meeussen2011long} have deployed a robot for 13 continuous days in an office environment, while exploring ways for improving the robot robustness by identifying failures and recoveries (including asking for help to humans). In \cite{biswas20161} a fleet of four CoBots reached 1,000 km of overall autonomous navigation. The robots were able to seek human assistance to perform manipulation tasks (the robots did not have arms) and send emails to developers in case of lack of human response. 
Similarly, in a deployment with the SCITOS G5 robot, that travelled more than 160 km within the STRANDS project \cite{hawes2017strands}, 
% Even though the robot was able of being autonomous for most of the time, 
the authors report the need for a way to manage failures and to have a better understanding of human activities. Hanheide et al. \cite{hanheide2017and} propose a spatio-temporal model to learn when, where and how users interacted with the robot info-terminal during a long-term deployment. They found they could improve the efficiency and usefulness of the system by proposing the right content at the right time and place.
Building from these works, we designed a system that is able to interact autonomously in a public environment for years while learning from the interactions with humans.
% \cite{rosenthal2012someone} presents a model that exploits the availability of humans in the environment by planning path that increase the likelihood of encountering of an active helper.
 % that includes the availability of humans in the environment, and
 % demonstrate how a navigation planner can incorporate this information to plan paths that increase the likelihood that a robot can find an available helper when it needs one.
A survey on long-term interaction between users and robots \cite{leite2013social} raises the issue that memory and adaptation remains nearly unexplored in the field.
% : survey that discussed several studies on long-term interaction between users and social robots. Among others, one of the key issues drawn from their analysis is that memory and adaptation remains nearly unexplored.
Similarly, Kunze et al. \cite{kunze2018artificial} explores the state of the art on Artificial Intelligence (AI) techniques for long-term autonomy
recognising that interactions in this context can be exploited to improve a robot behaviour.
% asserting that a major future challenge is that of integrating human interactions in the robot system to allow improv/ing its knowledge in unforeseen situations.
With the learning algorithm described in this work, we lay the foundations for a robot framework that allows
exploiting the human feedback during interaction in order to optimise its social abilities.

% In this project, we plan to exploit the human feedback during interaction to optimize the robot social abilities. \todo{really?}

 % address these issues by enabling the robot with the ability to directly optimize the users engagement during interaction and of learning to improve navigation robustness from users demonstrations.

Previous works featured a robot deployed in a museum environment. The robot Rhino \cite{burgard1998interactive} was deployed in a museum in Germany for 6 days guiding hundreds of visitors. At the time, the main issues and the focuses of the work were navigation and obstacle avoidance.
The Minerva robot \cite{thrun1999minerva} traversed more than 44 km and interacted with more than 50k people. Moreover, it was able to display mood (i.e. happy or angry) and, more importantly for our work, used an RL approach to learn the best actions to engage visitors.
In \cite{nourbakhsh2003mobot} four robots were deployed over five years. Focusing on interactivity and education they learned that long and non-interactive presentations are guaranteed to drive audience away.

%PB: this paragraph could be significantly improved: it is intended just to provide a brief summary of this section
In the present work, building on our ongoing long-term study of a museum robot \cite{del2019lindsey}, we plan to address some of these issues by enabling the robot with the ability to directly optimise the users' engagement during the interaction through RL. %and of learning to improve navigation robustness from users demonstrations.
% The first stage in doing so is characterising the engagement with the robot.

\subsection{Behavioural Adaptation in Social Settings}
% Previous work has shown that it is possible to influence the human engagement level during a human-robot interaction by employing different robot behaviors.
Previous work has shown how it is possible incorporate the users feedback, with a focus here on user engagement, for modifying the robot behaviour or influencing the user own engagement.
Sidner et al. \cite{sidner2005explorations} explored how the use of gazing and gestures affects positively the user perception of the robot, increasing their engagement. 
% reporting positive effect in terms of engagement.
Similarly, Holroyd \cite{holroyd2011generating} defines policies with the goal of increasing user engagement and shows that the robot equipped with these policies is perceived to be more human-like, to behave more fluently and that users reciprocate more robot cues.
% Lee et al. \cite{lee2019bayesian} proposes a computational framework to non-verbal communication in the context of storytelling human-robot settings.
Recent works aimed at learning these social behaviors typically use (Deep) RL techniques to exploit the real-world interaction experiences a robot can collect. Qureshi et al. \cite{qureshi2016robot, qureshi2017show} proposed end-to-end models to teach a robot the most appropriate action for approaching humans and starting an interaction. The reward signal was triggered by successful/unsuccessful handshakes.
% Positive and negative rewards were triggered by a sensor on the robot's hand whenever a successful/unsuccessful handshake was detected.
 % , it  has  been  proven  that,  through  machine  learning,  and  in particular with deep and reinforcement learning algorithms, it is possible to make robot gain social intelligence.
% works of hishiguro on deep RL on real robot...
Lathuili\`ere et al. \cite{lathuiliere2018deep} uses Deep RL to learn a gaze policy from an intrinsic reward function based on the audiovisual position of people with respect to the robot camera field of view. Gao et al. \cite{gaolearning} learns a robot policy for approaching groups of people by maximising a group formation score and minimising the displacement of other participants in the group when the robot approaches.
Also in a museum context, Meng et al. \cite{Meng2020} propose an RL approach where they use the users' engagement during group interactions with an interactive sculpture as the reward to learn engaging interactive behaviours.

% our take on this
In this work, we use a state-of-the-art RL approach \cite{Azar2017MinimaxRB} in order to improve our robot's social behavior, in particular its choice of actions during the guided tours. We employ the human engagement level
% , provided by our engagement prediction model above,
as a robot internal reward to maximise. 

 % for learning appropriate approaching behaviors. While both the approaches use an LSTM network to represent the policy, the first uses standard techniques to extract the feature state and the second uses features learned from a deep network.

% \cite{thomaz2016computational} "Hence, another challenge for social robots is to detect the human’s interaction engagement, both initially and in an ongoing fashion. Ideally this should be framed as a more generic recognition problem than that of recognizing a particular gesture or activity. This way the person does not have to make a specific gesture or use a speech command to initiate an interaction."

\section{PRELIMINARIES}
% here talk about

\subsection{Lindsey the Robot}
% Describe the robot used
Lindsey, the robot used in this long-term study, is a Scitos G5 robot manufactured by MetraLabs GmbH. In order to sense the environment, the robot has a laser scanner with 270$^{\circ}$ scan angle on its base and an Asus xtion depth camera mounted on a pan-tilt unit above his head. The interactions with the visitors are mediated through a touch screen, two speakers, a microphone and a head with two eyes that can move with five degrees of freedom to provide human-like expressions. To ensure safe operations in public environments the robot is equipped with an array of bumpers around the circular base with sensors to detect collisions and two easily reachable emergency buttons that, when activated, cuts the power to the motors.

The software framework is based on ROS and uses STRANDS project \cite{hawes2017strands} core modules for topological navigation, people tracking, task scheduling and data collection.

\subsection{The Museum Scenario} 
% Describe the museum and the tours that the robot does
% the fact that people can do whatever they want, not instructed

\begin{figure}[t]
    \centering
    \includegraphics[width=\columnwidth]{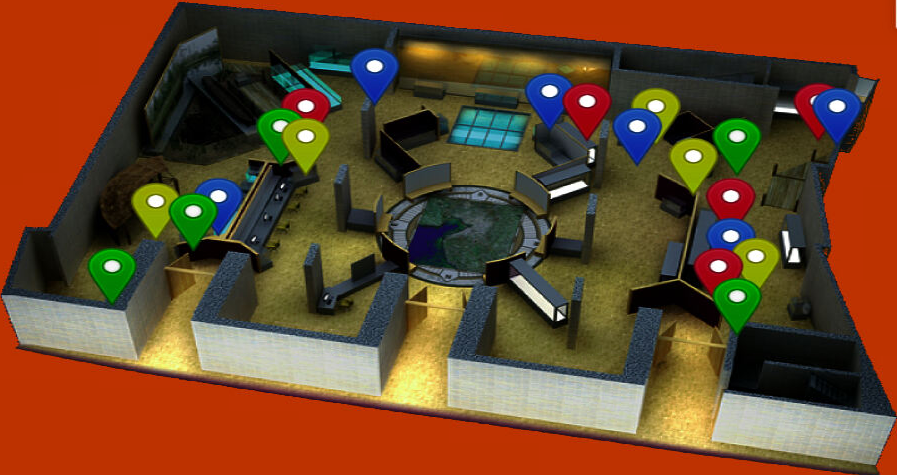}
    % \caption{Caption}
    % \label{fig:my_label}
    \caption{A representation of the archaeological gallery with markers to identify the position of items in the robot tours.}
    \label{fig:arc_map}
\end{figure}

\begin{figure}
    \centering
    \includegraphics[width=\columnwidth]{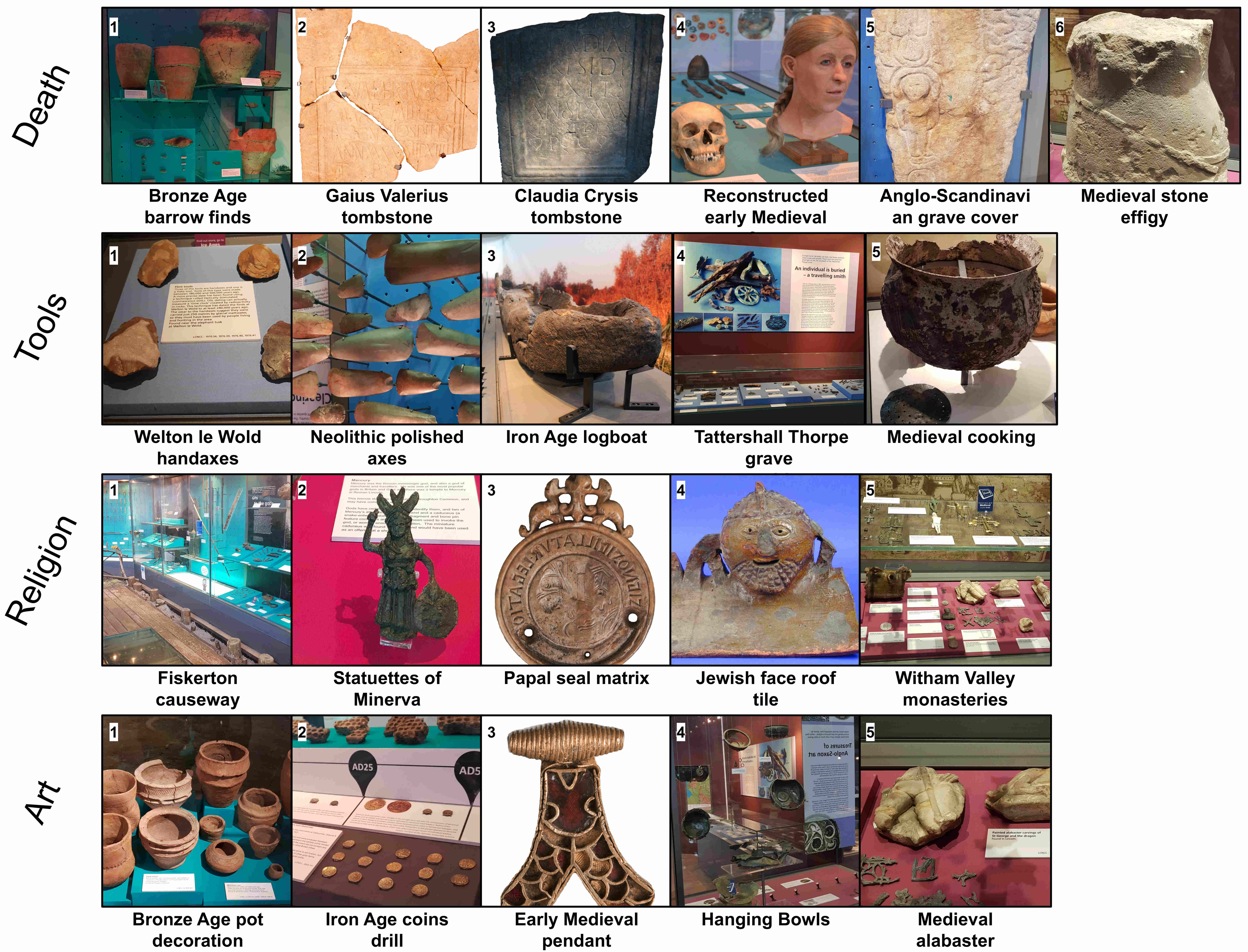}
    \caption{The sequence of items in each robot tour.}
    \label{fig:robot_tours}
\end{figure}

The scenario of this project is The Collection museum\footnote{\href{https://www.thecollectionmuseum.com/robot-at-the-collection}{https://www.thecollectionmuseum.com/robot-at-the-collection}} in Lincoln, UK. The museum is freely accessible to all members of the public 5 days a week, although it used to be open 7 days a week before the COVID lockdown, from 10AM to 4PM. Lindsey, the robot, is deployed as a tour guide in the archaeological section of the museum which displays local findings dating from the stone age to the early modern age.
Working with the museum's staff four different guided tours have been devised for the robot. Figure \ref{fig:arc_map} shows the position of the items visited by the robot during the tours. Different color markers correspond to items in different tours. Figure \ref{fig:robot_tours} depicts the items shown in each tour. 

The robot is free to roam around during idle times in search for people to interact with.
When people enter the gallery and go close to the robot they are greeted by it. Using the robot's touchscreen they can choose to start one of the tours or to being guided to the location of a specific item and receiving a description of it. For the purpose of this study only the guided tour interactions are considered.
The museum visitors are not instructed how to behave with the robot and what to expect during the interaction, therefore the interactions themselves are unstructured with users being free to interrupt the tours or to just leave at any moment.
Moreover, both the robot and the users move in the environment during the guided tours making the robot perception of users from the on-board cameras often limited.

\subsection{Analysis of the Robot Tours}
In a previous analysis of the interactions with our tour guide robot \cite{del2019lindsey} we report that Lindsey has been very popular among the museum's visitors, totalling more than 2300 guided tours in just 103 days of operations.
However, observing the duration of these tours one realises that most interactions are actually very short and that only about 18\% of them are completed by the user.
These observations evidences the need for adaptation and motivate us to study, with the present work, how the users' feedback can be taken into account in such long-term scenario to improve the robot ability to maintain engagement over time.

% \todo[inline]{here refer to the RO-MAN paper about the analysis of the operations. Highlight the low engagement rate for sustained interaction and therefore the need for learning and adaptation.}

\section{METHODOLOGY}
In this study we combine together several methodologies in order to have a unified framework that allows the robot to explore and learn online without the need of having separate phases for data collection and learning.

\subsection{States \& Actions Specification}
% How is the robot behaviour defined (PNP)
% and how this allow adaptation
% also talk how there are manual constraints in the plan to avoid executing nonsense.

The robot behaviours are specified as conditional plans using the Petri net plans (PNP) formalism \cite{ziparo2008petri}.
The formalism facilitates action abstraction a reusability in the specification of plans, moreover it allows to monitor the execution and deal with failure situations during execution with \emph{Execution Rules} (ER)~\cite{Iocchi2016Pratical}. 
Each action used in this study is a conditional sequence of lower-level actions with ERs for allowing robust execution. 
More information about how robustness during autonomous operations is assured can be found in our previous work~\cite{del2019lindsey}.
The duration of each action can vary from seconds to minutes and it is not known in advance, with some requiring navigating some distance around the museum or describing items with different amount of words. Moreover, each action is interruptible and can potentially terminate the tour before the end, for example in case of failure or users stopping the guided tour.   

\begin{table}[b]
\caption{Action space with constraints over possible action successors. The successors in {\color[HTML]{656565}grey} are conditional on that action not being executed already in the tour and those \underline{underlined} are only valid for the \texttt{death} tour.}
\label{tab:actions}
\centering
\begin{tabular}{|c|l|l|}
\hline
\textbf{Id} & \multicolumn{1}{c|}{\textbf{Action}} & \multicolumn{1}{c|}{\textbf{Action Successors}} \\ \hline
0           & \texttt{doNothing}                         & 0,{\color[HTML]{656565}1}                          \\ \hline
1           & \texttt{describeTour}                         & 2,3,4,5,6,\underline{7}                              \\ \hline
2           & \texttt{gotoExhibit\_1}                       & 8                                        \\ \hline
3           & \texttt{gotoExhibit\_2}                       & 8                                        \\ \hline
4           & \texttt{gotoExhibit\_3}                       & 8                                        \\ \hline
5           & \texttt{gotoExhibit\_4}                       & 8                                        \\ \hline
6           & \texttt{gotoExhibit\_5}                       & 8                                        \\ \hline
7           & \texttt{gotoExhibit\_6}                       & 8                                        \\ \hline
8           & \texttt{describeExhibit}         & {\color[HTML]{656565}2,3,4,5,6,\underline{7}},9,10 \\ \hline
9           & \texttt{describeMoreExhibit}     & {\color[HTML]{656565}2,3,4,5,6,\underline{7}},10\\ \hline
10          & \texttt{endTour}                              & 0                                         \\ \hline
\end{tabular}
\end{table}

\begin{figure*}[h]
    \centering
    \begin{subfigure}[b]{0.32\textwidth}
        \centering
        \includegraphics[width=\textwidth]{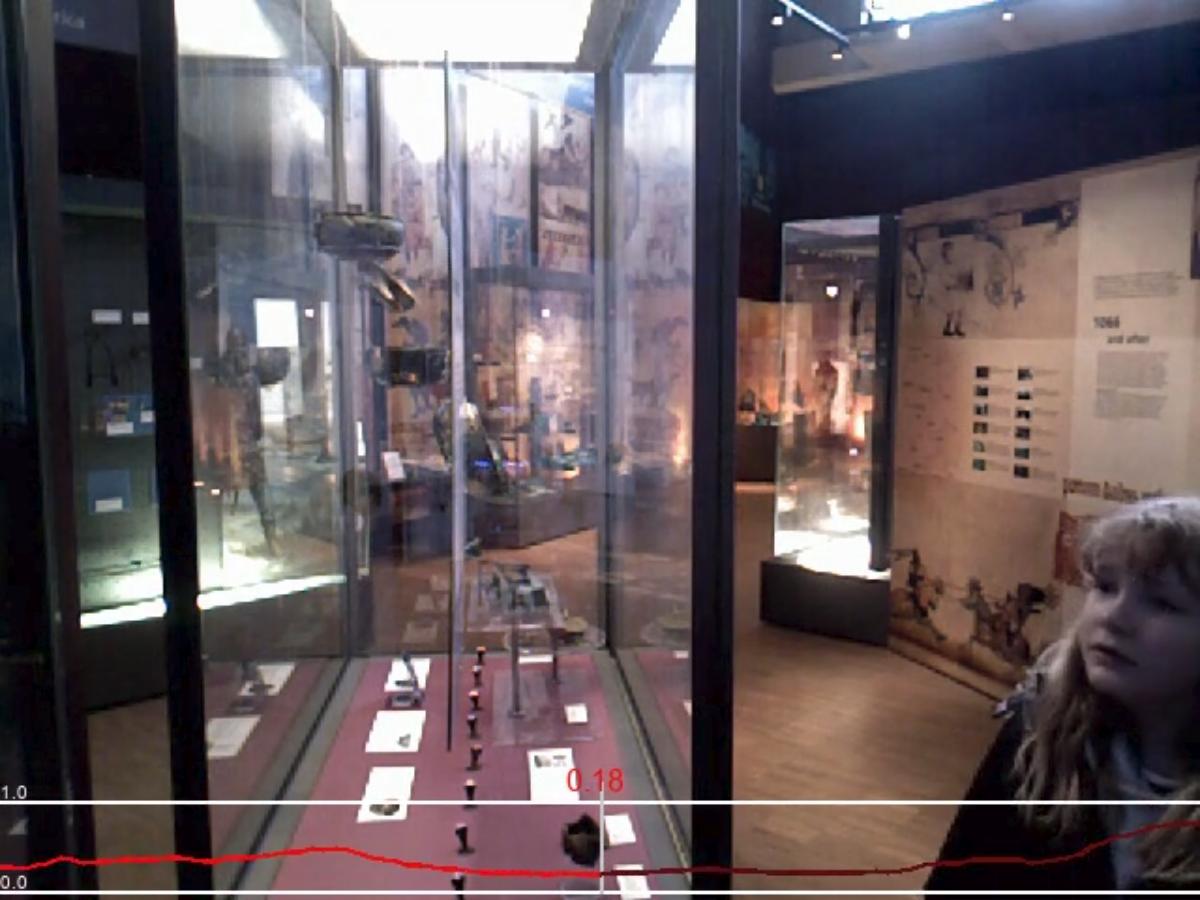}
    \end{subfigure}
    \hfill
    \begin{subfigure}[b]{0.32\textwidth}
        \centering
        \includegraphics[width=\textwidth]{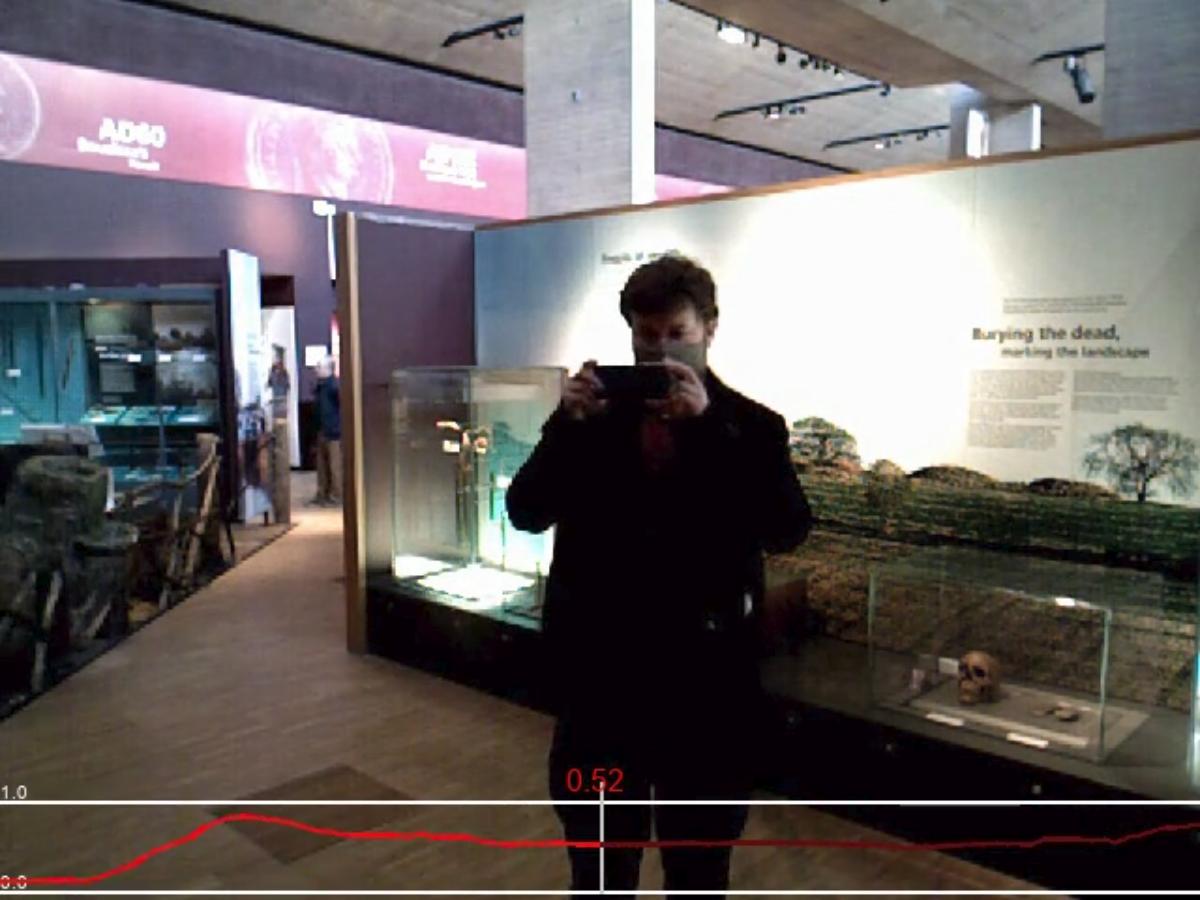}
    \end{subfigure}
    \hfill
    \begin{subfigure}[b]{0.32\textwidth}
        \centering
        \includegraphics[width=\textwidth]{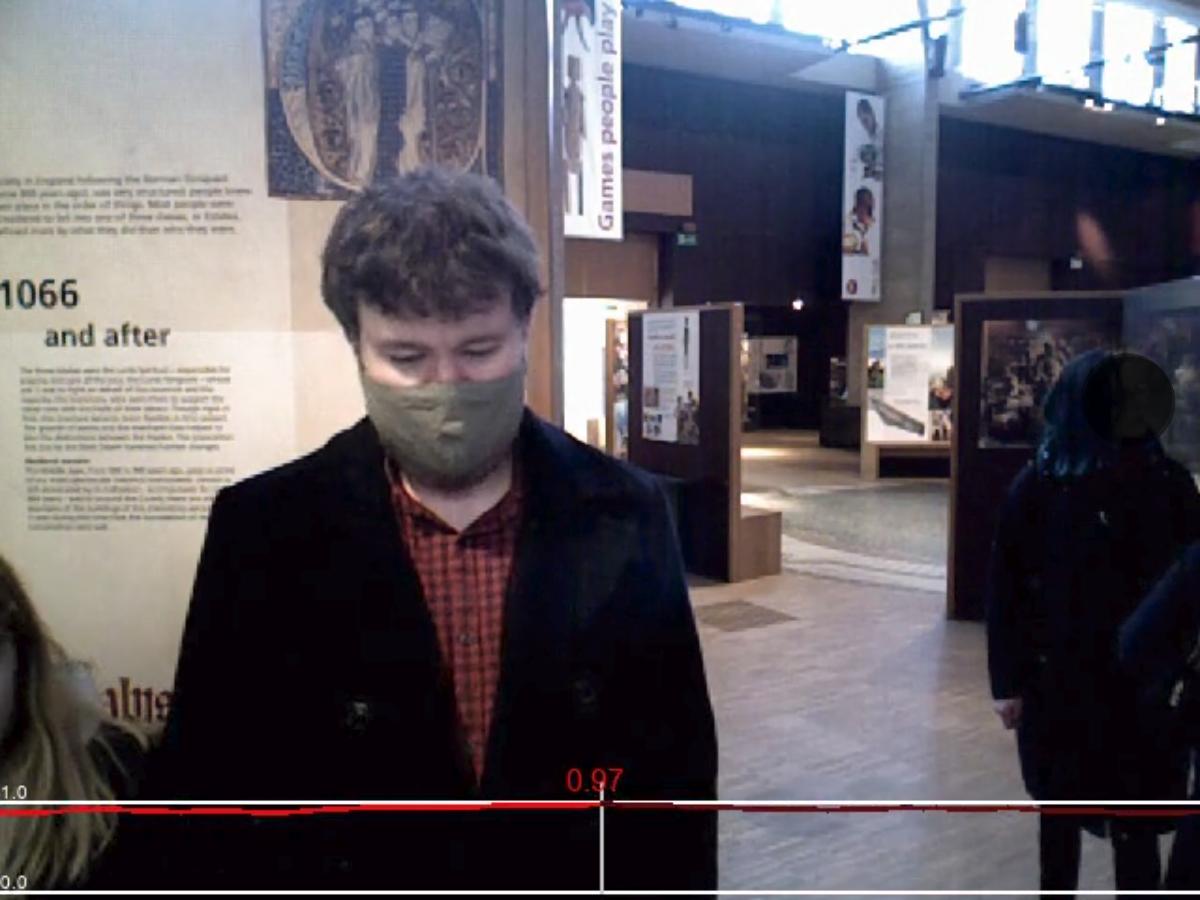}
    \end{subfigure}
    \caption{Examples of continuous engagement predictions from the robot head camera using model from \cite{del2020you}. Engagement prediction value is \texttt{LOW} in the left frame (the girl is looking elsewhere), \texttt{MEDIUM} in the center (man is taking a picture of the robot) and \texttt{HIGH} on the right (man looking at robot screen).  Written consent was obtained from the users for taking and reusing these pictures.}
    \label{fig:engagement_predictions}
\end{figure*}

% For our learning framework we select a subset of all the implemented actions,  which fall at a certain abstraction level that.
The scenario requires that certain actions are performed exclusively in a specific sequence. For example, it would be wrong to start describing an exhibit's item before going to its location or describing the theme tour only at the end of the interaction.
Moreover, the \texttt{gotoExhibit\_*} actions cannot be executed more than once because it wouldn't make sense to bring people to the same item in the museum multiple times during a tour. 
To enforce these constraints at execution time, we created subsets of all possible actions $C_s \subset A$ $\forall$ $s \in S$ from which the policy can choose the next action.  
The action space with the associated successive actions constraints is showed in Table \ref{tab:actions}.

The state vector used for learning is composed as follows
$$s = \langle v_1, v_2, v_3, v_4, v_5, v_6, n, a_p, e, t \rangle$$
where $v_k \in {0, 1}$ for $k=1,...,6$ indicates whether the $k^{th}$ item in the tour has been visited, $n \in \{\texttt{death}, \texttt{tools}, \texttt{religion}, \texttt{art}\}$ is the name of the tour, $a_p \in A$ is the previous action, $e \in \{\texttt{LOW}, \texttt{MEDIUM}, \texttt{HIGH}\}$ is the average engagement level during the execution of $a_p$, and $t \in \{\texttt{none}, \texttt{ended}, \texttt{stopped}, \texttt{abandoned}\}$ indicates the terminal state.

\subsection{Engagement Model}
\label{sec:eng_model}

% How the engagement model works and what it provides
The engagement of users who interact with Lindsey at the museum is detected using the engagement model presented in \cite{del2020you}, which runs in real-time on the robot's GPU.
The model proposed comprises and end-to-end regression model, taking the camera feed as input and providing a holistic engagement measure in the interval [0,1] as output for each image frame. It has been validated on HRI data sets as an accurate measure of engagement.
Figure \ref{fig:engagement_predictions} depicts 3 frames of a scene during  a tour with the continuous engagement signal provided by the model.
For the purpose of this work, we want to use the engagement of users as an evaluation of the robot behaviours to drive learning. 
Therefore, in our RL scenario the reward observed at each state is derived from the engagement collected during the execution of the last action.
The engagement scalar values are provided by the model at a frequency of about 1 Hz, they are then averaged over the entire duration of the action and discretised in the variables \texttt{LOW}, for values in $(0, \frac{1}{3}]$, \texttt{MEDIUM}, for values in $(\frac{1}{3}, \frac{2}{3}]$,  or \texttt{HIGH}, for values in $(\frac{2}{3}, 1]$.
Finally, the reward function, given the state $s$, is 
$$r(s) \sim \begin{cases}
               \mathcal{U}(0, \frac{1}{3}) & \text{if $e =$ \texttt{LOW}},\\
               \mathcal{U}(\frac{1}{3}, \frac{2}{3}) & \text{if $e =$ \texttt{MEDIUM}},\\
               \mathcal{U}(\frac{2}{3}, 1) & \text{if $e =$ \texttt{HIGH}}.
            \end{cases}$$
% The engagement is provided by the engagement model presented in \cite{del2020you} which runs in real-time on the robot's GPU during the operations. 
% The engagement value rate is provided at a frequency of about 1 Hz. At any moment, given a frame $x_k$ from the robot's upper camera, the engagement model provides an engagement value of the scene $e_k = \mathrm{M}_e(x_k)$.

% explain that one could replace this by giving positive rew at the end of episode, or giving 1 at each step given that the longer the tour the better for us. But, by giving the detected engagement we encourage planning actions that lead to higher future engagement.
We provide the scalar value of engagement as reward in order to guide the learned policy toward actions that are expected to elicit higher future engagement in the users. Also, given that $r(s) > 0$ $\forall \,s \in S$ there is an implicit effect of favouring longer tours. 

\subsection{Behaviour Adaptation with UCBVI}~\label{sec:ucbvi}
% The UCBVI algorithm used; how the learning is done in real-time; talk about the sparse update of the model to be faster; talk about the exploration bonus.

%%%%% not sure this should go here
% \todo{maybe move this somewhere else}
When training an agent to learn in an unknown environment, the \emph{exploration-exploitation dilemma} quickly arises where the agent needs to decide if to keep exploring new actions or executing the one that has returned the highest rewards so far.  
This trade-off is particularly relevant in robotic applications, like ours, where the amount of exploration one can perform is costly in terms of time and resources and limited by the dynamics of the real world.
% , and where  one wants the robot to start exploiting the good actions it learned as soon as possible rather than finishing exploring the other .  
In this work, for online learning, we use 
% Upper-Confidence Bound Value Iteration (UCBVI) \cite{Azar2017MinimaxRB}, 
a Reinforcement Learning algorithm based on the ``optimism in the face of uncertainty'' principle. 
This principle prescribes that when the model of the environment is uncertain, one should consider the best possible world; if the model is correct, you have \emph{no regrets} (exploitation); otherwise, you have effectively learned something new about the world (exploration).
This general principle gives an almost optimal solution for the stochastic multi-armed bandit problem \cite{Bubeck2012} and episodic \cite{Azar2017MinimaxRB} and ergodic \cite{Jaksch2008NearoptimalRB} RL problems.
%%%%%

In this work, we are dealing with an episodic RL problem where every guided tour given by the robot is an episode. 
For improving the robot policy we use the Upper Confidence Bound Value Iteration (UCBVI) algorithm\cite{Azar2017MinimaxRB} which allows to balance the exploration of novel state-action pairs with the exploration of already explored ones. 
In out setting this is particularly important because we have a part of the state-action space which was explored very well during the initial deployment with the static policy. UCBVI gives a natural way of incorporating previously accumulated data without biasing the policy to the point that it doesn't explore new actions.
Algorithm \ref{alg:ucbvi} outlines our implementation of the algorithm. 
As in standard \emph{model based} Value Iteration, the algorithm is composed of 3 phases. First, the value function is computed from the model and an initial policy is generated, then the policy is used for acting in the environment while collecting a new episode, finally the model is improved based on the updated dataset of episodes. 
The loop repeats, in our case, for every new guided tour that is requested by the users.

UCBVI favours exploration of novel state-action pairs thanks to a bonus function that increases the $Q$-value for pairs that have been scarcely visited. The bonus decreases toward zero as we obtain more data and the $Q$-value tends to the real value.
Algorithm \ref{alg:bonus} shows the bonus, inspired by the Chernoff-Hoeffding’s bonus proposed in \cite{Azar2017MinimaxRB}.
The speed with which the bonus tends to zero is set by the free variable $\sigma$. For this work, it was set to $\sigma=20$ in order to allow a relatively fast exploration of new state-action pairs in our real-world scenario, where obtaining new episodes is difficult, but it can be set lower in situations where performing exploration is easier, for example simulations.  
% that was chosen for this work following empirical experimentation.

% The reward function is known and corresponds to the averaged engagement value detected during the last action $r_t = \frac{\sum^k_K \mathrm{M}_e(x_k)}{\mid K \mid}$. 
% The transition function is unknown and must be learned from exploration data.
% \todo[inline]{In Algo 1 some forall are declared in a lazy way...is that okay?}
\begin{algorithm}[t] \caption{Online learning algorithm}\label{alg:ucbvi}
\KwData{$S$, $A$, $H$, $C_s$ $\forall$ $s \in S$}
$Q_h(s,a) \gets 0$ $\forall \,\,(s,a) \in S \times A$ and $h=1,...,H$\;
$N_h(s,a,s') \gets 0$ $\forall \,\,(s,a,s') \in S\times A\times S$ and $h=1,...,H$\;
$\mathbb{D} \gets \emptyset$\;
\ForEach(){episode}{
    \tcp{Policy optimisation}
    \For{$h=H,...,1$}{
        \For{$(s,a) \in S \times A$}{
            $b_h(s,a) \gets \texttt{bonus}(h,s,a)$\;
            $Q_h(s,a) \gets r(s) + b_h(s,a) + \mathbf{E}_{s'\sim \hat{p}_h(\cdot\mid s,a)}[V_{h+1}(s')]$\;
            $V_h(s) \gets \min{\{H-h, \max_{a\in A}Q_h(s,a)\}}$\;
        }
    }
    $\pi_h(s) \gets \text{arg\,max}_{a\in C_s}Q_h(s,a)$ $\forall \,\,s \in S$\;
    \tcp{Policy execution}
    Observe $s_1$\;
    \For{$h=1,...,H$}{
        Execute $a_h \gets \pi_h(s_h)$\;
        Observe $r_h$ and $s_{h+1}$\;
        Add $(s_h,a_h,s_{h+1})$ to $\mathbb{D}$
    }
    % Add trajectory $(s_h,a_h)^{H-1}_{h=0}$ to $\mathbb{D}$\;
    \tcp{Model update}
    $N_h(s,a,s') \gets \mid\{(s_h,a_h,s_{h+1})\in\mathbb{D}:(s_h,a_h,s_{h+1})=(s,a,s')\}\mid$ $\forall \,\, (s,a,s',h)$\;
    % $N_h(s,a) \gets \sum{1_{\mathbb{D}}(s,a)_h}$ $\forall \,\,(s,a) \in S\times A$ and $h=1,...,H$\;
    $\hat{p}_h(s'\mid s,a) \gets \frac{N_h(s,a,s')}{\mid N_h(s,a,\cdot)\mid}$ $\forall \,\,(s,a,s',h)$\;
}
\end{algorithm}
\begin{algorithm}[t] \caption{\texttt{bonus}}\label{alg:bonus}
\KwData{$N_h(s,a,s')$, $H$}
$N \gets \mid N_h(s,a,\cdot)\mid$\;
$b \gets \sqrt{\frac{1}{N}} + \frac{H-h}{\sigma\cdot N}$\;
\Return $\min\{b, H-h\}$
\end{algorithm}

% not sure if this should be here or not
In our implementation, is important to notice that the $Q$-function defined favours the exploration of poorly explored areas over other areas that have never been observed.
The effect is that the policy sticks to choosing the same, poorly explored, actions over consecutive episodes and starts exploring a new one only after a certain number of visitations have been done, rather than continuously selecting different actions to explore.  
This effect, which we found quite useful to maintain a more consistent behaviour over time for our on-line application, could be eliminated by \emph{optimistically} initialising $Q_h(s,a) = H-h$ $\forall \,\,(s,a) \in S \times A$ and $h=1,...,H$.

\section{EXPERIMENTAL VALIDATION}

In order to validate the hypothesis that optimising a social robot policy to maximise the user engagement leads to more sustained human-robot interactions we have performed a long-term study in the museum with our tour guide robot Lindsey. 
Since 2019, Lindsey has been delivering the guided tours depicted in Figure \ref{fig:robot_tours} to the visitors in a "static" way. With its static policy, during each tour the robot would guide the users to the exhibits always in the same order and the choice of whether to give a more detailed description about each item was left to the user.
Motivated by the fact that users' displaying different perceived engagement have different willingness to continue the interaction \cite{peters2005model}, as shown also by our data in Figure \ref{fig:prob_continue}(a) where for example a perceived low engagement at the first stop of the tour means a 20\% decrease in chances of users continuing the interaction, we attempt to employ the users engagement value as a factor to select different actions during the interactions. 

In our learning scenario, we allow the policy to: 1) choose freely the order of items visited in the tours; and 2) decide whether to provide additional information about each specific item, rather than asking the user.
The model of our scenario, namely the visitation frequency table $N_h(s,a,s')$, is initialised with the transitions from the thousands of episodes collected with the static policy. 
During the learning phase, our \emph{optimistic} learning algorithm implemented takes care of directing the exploration toward the other less explored areas.
The reward function is computed on-line from the engagement values detected by our engagement model, as described in Section \ref{sec:eng_model}.
The robot deployment, interrupted in 2020 because of COVID lockdown, was resumed in December 2021 and the learning phase began. During the following 8 weeks the robot behaviour was driven by our online learning algorithm which explored the different actions it could now take, while exploiting the model to increase users' willingness to continue.

In order to assess more properly the effect of our learned policy on the tour success and on users' engagement we perform a 2 week verification phase, during which the robot behaviour is again driven by the original static policy. This validation was necessary to observe whether the effect over our results during the learning phase was actually caused by our learned policy or by other spurious effects, like people willing to spend more time at the museum after COVID lockdowns or face masks wearing.

\subsection{Performances of Learned Policy}~\label{sec:performances}
% Here show the average number of stops over time and the probability of successful tours over time
% Compare with data over the last two weeks for validation
% Show how the tour has changed

\begin{figure}[t]
    \centering
    \includegraphics[width=\columnwidth]{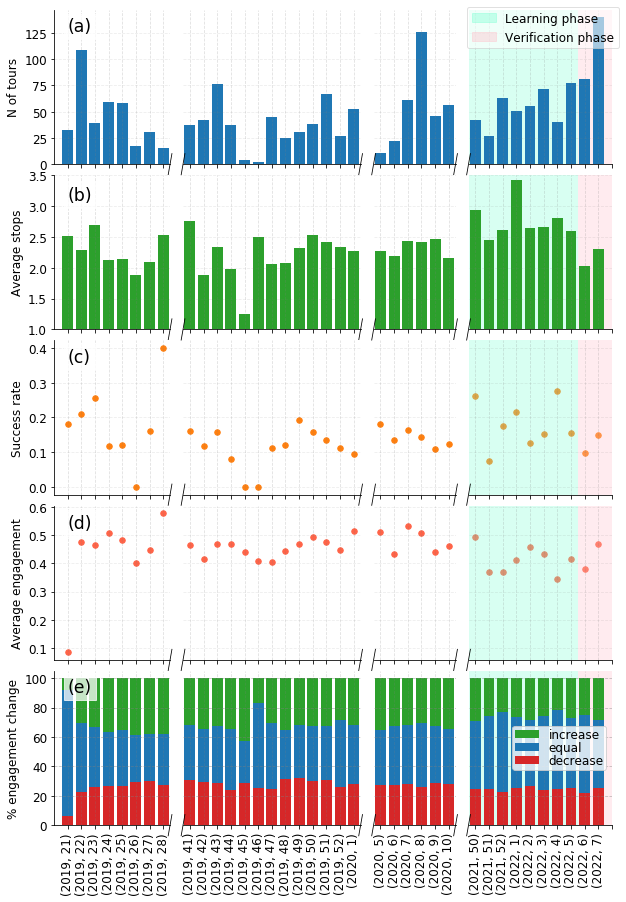}
    \caption{(a) Number of guided tours delivered per week, (b) average number of stops visited in  the tours, (c) rate of tours completed to the end, (d) average scalar engagement level during tours (e) average change in engagement binned value (i.e. from \texttt{HIGH}, \texttt{MEDIUM}, \texttt{LOW}) between each pair of consecutive actions. Missing time ranges are caused by the robot not being operational or museum closures. }
    \label{fig:tour_success}
\end{figure}

In this section we report the results of our evaluation to assess the performances  of the learned robot behaviour. 
In Figure \ref{fig:tour_success} we report the evolution of different metrics over the entire deployment (omitting the periods during which the robot was not operational). 
The values shown are collected and averaged per-week with the number of tours for each data point reported in Figure \ref{fig:tour_success}(a) for significance.
During the whole deployment period the museum scenario did not change in any major way to affect the robot tour guides. Similarly, the robot structure and touch screen interface had only minor interventions like the replacement of an RGB-D sensor.
% The results highlighted by the light blue band correspond to those obtained while the learning algorithm is generating the robot policy, the rest of the results are collected from the static policy. 
Data reported for the weeks from (2019, 21) to (2020, 10) and after (2022, 5) are results of the "static" tours, while from (2021, 50) to (2022, 5) are caused by the learned policy.

In Figure \ref{fig:tour_success}(b) we report the average number of stops visited  (exhibit described) in the tours. During the learning phase we observe an increase of 22.8\% over the previous period, with the verification phase bringing back the values in line with the previous data for the static tour.
Similar results are obtained for the tour success rate, i.e. the rate of tours that terminate after visiting all the stops, with an overall increase of over 30\% with respect to the static tour as reported in Figure \ref{fig:tour_success}(c).
Notice that the nominal number of items is 6 for the \texttt{Death} tour and 5 for the rest, therefore a lower number of items means that the users have stopped or abandoned the interaction before the end.
These results together suggest that our learned policy can indeed lead users to keep a more sustained engagement during the interactions in the museum, confirming the hypotheses which motivates this work.
In addition, we study the average users' engagement detected by our model during the whole duration of the tours and the average change in engagement over each consecutive actions executed. The plots in Figure \ref{fig:tour_success}(d)-(e) suggest that during the learning phase there was a general decrease of detected engagement with respect to the previous and the verification periods. Also, we can observe that the learned policy had the effect of maintaining the users' engagement stable over time almost 50\% of the times, differently from the previous period where there was an equal chance of increasing, decreasing or maintaining stable the engagement.

\begin{figure}
    \centering
    \includegraphics[width=\columnwidth]{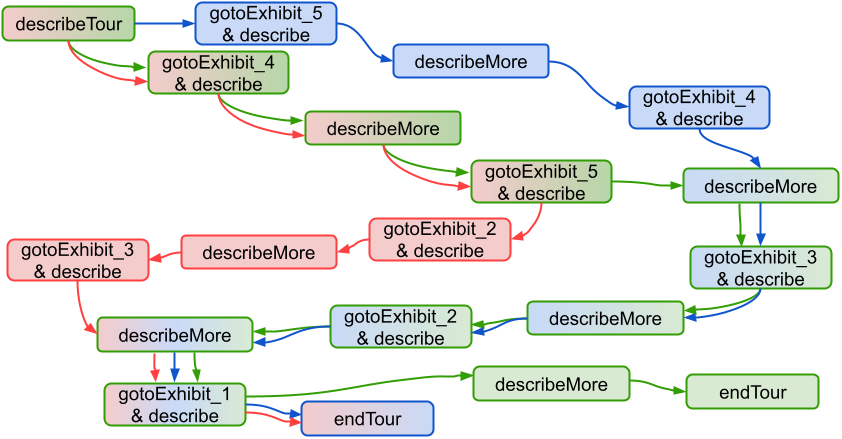}
    \caption{Different actions chosen by the learned policy for the tour \texttt{art} at different levels of engagement. Engagement values are red for \texttt{LOW}, blue for \texttt{MEDIUM} and green for \texttt{HIGH}. }
    \label{fig:art_learned_tour}
\end{figure}

\subsection{Analysis of Learned Tours}~\label{sec:tour_analysis}

\begin{figure}[t]
    \centering
    \begin{subfigure}{0.48\columnwidth}
        \centering
        \includegraphics[width=\textwidth]{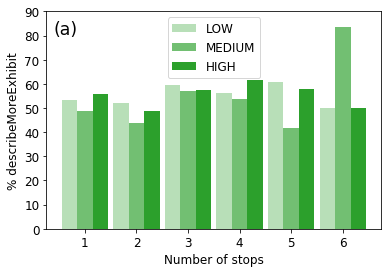}
    \end{subfigure}
    \begin{subfigure}{0.48\columnwidth}
        \centering
        \includegraphics[width=\textwidth]{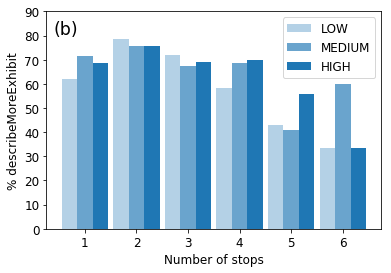}
    \end{subfigure}
    \caption{Percentage of times additional information is provided to the user at different steps of the tour and for different values of engagement. (a) For static tour, the additional information is requested by the user; (b) the learned policy decides when to give more information. }
    \label{fig:prob_descmore}
\end{figure}

In this section we analyse how the guided tour has changed after the learning phase, with respect to the original static tour.
One of the effect of allowing the learning algorithm to explore different actions was that now the robot could scramble the sequence of the items in the tour, guiding people to different places in order. In Figure \ref{fig:art_learned_tour} we show, as an example of this change, how the final learned policy would conduct the \texttt{art} tour at different level of engagements detected during execution. 
The figure evidences how different users' engagement levels generates different tours. Additionally, the different engagements had the effect of changing the amount of information given to the users at each exhibit. In Figure \ref{fig:prob_descmore} we report how many times an additional description is given to user, by performing the \texttt{describeMoreExhibit} action, for each stop number of the tour at different engagement levels. We observe that the robot learned to give it more often at the beginning of the tour, while for later steps only when the engagement is \texttt{MEDIUM} or \texttt{HIGH}.

\begin{figure}[t]
    \centering
    \begin{subfigure}{0.48\columnwidth}
        \centering
        \includegraphics[width=\textwidth]{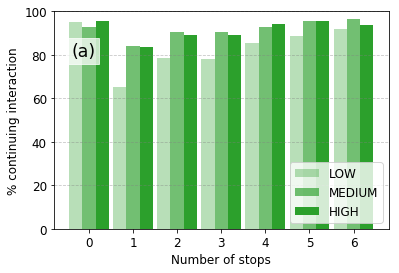}
    \end{subfigure}
    \begin{subfigure}{0.48\columnwidth}
        \centering
        \includegraphics[width=\textwidth]{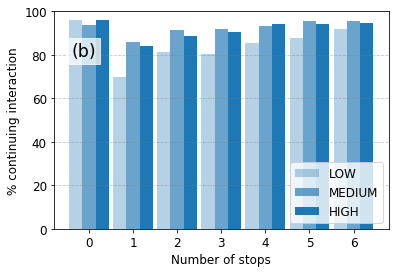}
    \end{subfigure}
    \caption{Percentage of times the tour is continued, rather than stopped or abandoned, at different number of items visited in the tour and for different values of engagement. (a) Static tour, (b) learned tour. Stop 0 corresponds to the \texttt{describeTour} action, before having reached any exhibit.}
    \label{fig:prob_continue}
\end{figure}

Finally, to verify whether a different engagement level correlates with a different willingness to stay in the interaction we study how many times a tour is stopped or abandoned varying the engagement value at each tour stop. The results in Figure \ref{fig:prob_continue} shows that users are more probable to disengage in the earlier stops, after the initial tour description, and when they show an already \texttt{LOW} engagement level, consistent with our intuition that poorly engaged users are less willing to continue an interaction in the first place. 
We observe an overall decrease in disengagement when using the learned policy compared to the static policy, but no other significant difference is present between the two conditions.

\subsection{State-Action Exploration}
% Show data about the exploration
% Show the simulation of future exploration and explain that the more we explore the more is difficult to encounter the remaining states 
\begin{figure}[t]
    \centering
    \includegraphics[width=\columnwidth]{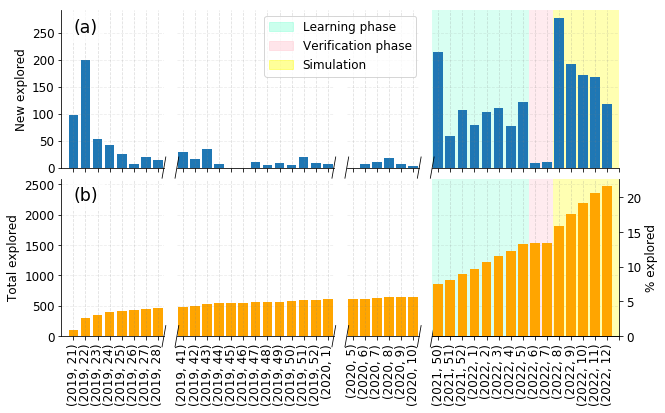}
    \caption{Exploration of the state-action space of the policy per week. (a) Number of newly explored state-action pairs, (b) cumulative exploration.}
    \label{fig:exploration}
\end{figure}

The UCBVI algorithm used in this work gives a simple yet efficient way of exploring new unseen states while exploiting the best performing actions, as described in Section \ref{sec:ucbvi}. In this section we study how the algorithm has effectively explored the state-action space in our scenario.

Figure \ref{fig:exploration} shows the amount of exploration that the robot has done, reported per week. As per our expectations, when using the static tour little to no exploration is performed, given that the robot is always choosing the same actions at the same point in  the tour. 
Note that in this condition a little amount of exploration is still ongoing, but it is entirely governed by the users' behaviour with their choices during the tour, i.e. asking for more information or not, and by manifesting different levels of engagement. 
The adoption of our learning algorithm has brought an initial week of high exploration of many new state-action transitions that were previously forbidden and kept a steady exploration rate in the subsequent weeks. 
The exploration is smaller after the initial period given that the policy starts to exploit the new actions that have resulted being most promising in terms of future users' engagement, as evidenced also by the analysis of the learning outcomes in Sections \ref{sec:tour_analysis} and \ref{sec:performances}.

The exploration rate achieved as of the beginning of 2022 is only about 13\% of the entire state-action space. We expect that as the learning phase is resumed, after the verification period, with more exploration the robot will achieve higher performances in terms of being able to retain users interest over time in the tours. 
However, we hypothesise that the more we continue to act and explore, the more will be difficult to find new states to explore. As it was mentioned in the previous paragraph, some parts of the state-action space are not directly explorable by the robot but they can only be observed indirectly after the user behaviour allows so.
To validate this hypothesis, we have performed a simulation of the next 5 weeks into the future where we keep using and improving our learned policy and we simulate the next observed states by sampling uniformly from the set of all the possible that could be observed at each point in the tour. 
The simulated exploration, shown in Figure \ref{fig:exploration}, is an upper-bound of the empirical exploration we can obtain in real life because it does not simulate the possibility of \emph{stopping} or \emph{abandoning} the interaction, which would ultimately prevent exploring further into the tour sequence. However, the data shows that as the time passes the amount of states that can be explored by the algorithm decreases.

\section{DISCUSSION \& CONCLUSIONS}
% Discuss:
%   - why it was necessary to collect again data with a static policy (ie. there may be other factors influencing the users behaviour)
%   - The exploration is low at this point, but it doesn't matter as long as the behaviour has improved. COVID did not allow to collect more data. We expect that it will keep increasing as we explore more, because of how UCBVI works this must happen.
In this work, we proposed an Reinforcement Learning approach to in-situ behavioural adaptation for social robots. The framework exploit a long-term museum deployment of a tour guide robot for learning, by experience, what are the best actions the robot should perform during the guided tour to sustaining the users interaction with the robot for longer.
The user engagement is detected from the robot own camera in real-time to be used both as a part of the state, to plan actions that are more appropriate for the displayed engagement, and as a continuous reward to the robots actions, to encourage it executing actions that lead to eliciting higher future engagement.
With our experimental validations, we observe that the adaptation framework leads the robot to perform longer tours and the users to stop, or abandon, the interaction less frequently than before with a static policy.
Even though only a fraction of the entire state-action space has been explored, complete exploration cannot be practically achieved in our real-world scenario. 
The UCBVI algorithm implemented is capable of exploiting the most promising actions explored so far, hence maintaining good overall performances, while exploring new states and actions, as confirmed by our empirical results.
% we can be confident after the verification phase's results that our online learning framework enables our robot Lindsey to deliver situated interactions leading to more sustained human-robot interactions. 

Our study present results that span over a period of almost 3 years, including the COVID pandemic, which could very well encompass many factors influencing the users' behaviour over time.
% Therefore, a 2-week verification phase, which successfully verified that our improved performances are caused by the learning approach, was performed. 
% A possible effect detected
% The results about the detected engagement in Figure \ref{fig:tour_success}(d)-(e)
In fact, when reporting the users' detected engagement, we observe an overall decrease since the museum reopening after lockdown which could be caused by mask wearing affecting the ability of our model to detect engagement. However, the improved results of our learned policy over the static one, even confirmed by the two-weeks verification period, shows that the overall learning framework maximising detected engagement is able to produce more sustained human-robot interactions.

% The tour in Figure \ref{fig:art_learned_tour}, but also for the others, we notice that the learned policy kept a preference to visit items nearby in sequence event though the order is different. This possibly reflects the preference of the users to perform the tour in an order that entails navigating along the shortest path \colorbox{red}{measure the average distance navigated now and before}.

%   - This work does not address what if the people behaviour change over time. We would need to modify the UCBVI bonus to also take in consideration how much the latest samples deviate from the current Q estimates. 
A natural extension of this work is to study how our learning framework would allow the adaptation of the robot behaviour over time as people's preferences change. The UCBVI bonus of the present algorithm favours exploration of poorly explored areas, but it would not recognise whether the estimation of well explored areas is still relevant, or it needs more experience caused by a shift in users' behaviour. A general idea to integrate this would be to have the bonus factor in the divergence between the values of newly explored state-action pairs and their estimation from past data.

%   - Many more features could be added to the policy state-space but this would need approximation techniques (i.e. deep learning) to deal with the higher dimentionality. Cite recent papers on DEEP UCBVI. This would also pose challenges on the robot hardware (bigger GPU?)


%%%%%%%%%%%%%%%%%%%%%%%%%%%%%%%%%%%%%%%%%%%%%%%%%%%%%%%%%%%%%%%%%%%%%%%%%%%%%%%%

%%%%%%%%%%%%%%%%%%%%%%%%%%%%%%%%%%%%%%%%%%%%%%%%%%%%%%%%%%%%%%%%%%%%%%%%%%%%%%%%

%%%%%%%%%%%%%%%%%%%%%%%%%%%%%%%%%%%%%%%%%%%%%%%%%%%%%%%%%%%%%%%%%%%%%%%%%%%%%%%%

%%%%%%%%%%%%%%%%%%%%%%%%%%%%%%%%%%%%%%%%%%%%%%%%%%%%%%%%%%%%%%%%%%%%%%%%%%%%%%%%

\bibliography{biblio}

\begin{thebibliography}{10}
\providecommand{\url}[1]{#1}
\csname url@rmstyle\endcsname
\providecommand{\newblock}{\relax}
\providecommand{\bibinfo}[2]{#2}
\providecommand\BIBentrySTDinterwordspacing{\spaceskip=0pt\relax}
\providecommand\BIBentryALTinterwordstretchfactor{4}
\providecommand\BIBentryALTinterwordspacing{\spaceskip=\fontdimen2\font plus
\BIBentryALTinterwordstretchfactor\fontdimen3\font minus
  \fontdimen4\font\relax}
\providecommand\BIBforeignlanguage[2]{{%
\expandafter\ifx\csname l@#1\endcsname\relax
\typeout{** WARNING: IEEEtran.bst: No hyphenation pattern has been}%
\typeout{** loaded for the language `#1'. Using the pattern for}%
\typeout{** the default language instead.}%
\else
\language=\csname l@#1\endcsname
\fi
#2}}

\bibitem{del2019lindsey}
F.~Del~Duchetto, P.~Baxter, and M.~Hanheide, ``Lindsey the tour guide
  robot-usage patterns in a museum long-term deployment,'' in
  \emph{International Conference on Robot \& Human Interactive Communication
  (RO-MAN)}.\hskip 1em plus 0.5em minus 0.4em\relax IEEE, 2019.

\bibitem{del2020you}
F.~{Del Duchetto}, P.~Baxter, and M.~Hanheide, ``{Are You Still With Me?
  Continuous Engagement Assessment From a Robot's Point of View},''
  \emph{Frontiers in Robotics and AI}, vol.~7, sep 2020.

\bibitem{meeussen2011long}
W.~Meeussen, E.~Marder-Eppstein, K.~Watts, and B.~P. Gerkey, ``Long term
  autonomy in office environments,'' in \emph{ALONE Workshop, In Proceedings of
  Robotics: Science and Systems (RSS’11), Los Angeles, USA}, 2011.

\bibitem{biswas20161}
J.~Biswas and M.~Veloso, ``The 1,000-km challenge: Insights and quantitative
  and qualitative results,'' \emph{IEEE Intelligent Systems}, vol.~31, no.~3,
  pp. 86--96, 2016.

\bibitem{hawes2017strands}
N.~Hawes, C.~Burbridge, F.~Jovan, L.~Kunze, B.~Lacerda, L.~Mudrova, J.~Young,
  J.~Wyatt, D.~Hebesberger, T.~Kortner, \emph{et~al.}, ``The strands project:
  Long-term autonomy in everyday environments,'' \emph{IEEE Robotics \&
  Automation Magazine}, vol.~24, no.~3, pp. 146--156, 2017.

\bibitem{hanheide2017and}
M.~Hanheide, D.~Hebesberger, and T.~Krajn{\'\i}k, ``The when, where, and how:
  An adaptive robotic info-terminal for care home residents,'' in
  \emph{Proceedings of the 2017 ACM/IEEE International Conference on
  Human-Robot Interaction}.\hskip 1em plus 0.5em minus 0.4em\relax ACM, 2017,
  pp. 341--349.

\bibitem{leite2013social}
I.~Leite, C.~Martinho, and A.~Paiva, ``Social robots for long-term interaction:
  a survey,'' \emph{International Journal of Social Robotics}, vol.~5, no.~2,
  pp. 291--308, 2013.

\bibitem{kunze2018artificial}
L.~Kunze, N.~Hawes, T.~Duckett, M.~Hanheide, and T.~Krajn{\'\i}k, ``Artificial
  intelligence for long-term robot autonomy: A survey,'' \emph{IEEE Robotics
  and Automation Letters}, vol.~3, no.~4, pp. 4023--4030, 2018.

\bibitem{burgard1998interactive}
W.~Burgard, A.~B. Cremers, D.~Fox, D.~H{\"a}hnel, G.~Lakemeyer, D.~Schulz,
  W.~Steiner, and S.~Thrun, ``The interactive museum tour-guide robot,'' in
  \emph{Aaai/iaai}, 1998, pp. 11--18.

\bibitem{thrun1999minerva}
S.~Thrun, M.~Bennewitz, W.~Burgard, A.~B. Cremers, F.~Dellaert, D.~Fox,
  D.~Hahnel, C.~Rosenberg, N.~Roy, J.~Schulte, \emph{et~al.}, ``Minerva: A
  second-generation museum tour-guide robot,'' in \emph{Proceedings 1999 IEEE
  International Conference on Robotics and Automation (Cat. No. 99CH36288C)},
  vol.~3.\hskip 1em plus 0.5em minus 0.4em\relax IEEE, 1999.

\bibitem{nourbakhsh2003mobot}
I.~R. Nourbakhsh, C.~Kunz, and T.~Willeke, ``The mobot museum robot
  installations: A five year experiment,'' in \emph{Proceedings 2003 IEEE/RSJ
  International Conference on Intelligent Robots and Systems (IROS 2003)(Cat.
  No. 03CH37453)}, vol.~4.\hskip 1em plus 0.5em minus 0.4em\relax IEEE, 2003,
  pp. 3636--3641.

\bibitem{sidner2005explorations}
C.~L. Sidner, C.~Lee, C.~D. Kidd, N.~Lesh, and C.~Rich, ``Explorations in
  engagement for humans and robots,'' \emph{Artificial Intelligence}, vol. 166,
  no. 1-2, pp. 140--164, 2005.

\bibitem{holroyd2011generating}
A.~Holroyd, ``{Generating engagement behaviors in human-robot interaction},''
  Master's thesis, Worcester Polytechnic Institute, 2011.

\bibitem{qureshi2016robot}
A.~H. Qureshi, Y.~Nakamura, Y.~Yoshikawa, and H.~Ishiguro, ``Robot gains social
  intelligence through multimodal deep reinforcement learning,'' in \emph{2016
  IEEE-RAS 16th International Conference on Humanoid Robots (Humanoids)}.\hskip
  1em plus 0.5em minus 0.4em\relax IEEE, 2016, pp. 745--751.

\bibitem{qureshi2017show}
------, ``Show, attend and interact: Perceivable human-robot social interaction
  through neural attention q-network,'' in \emph{2017 IEEE International
  Conference on Robotics and Automation (ICRA)}.\hskip 1em plus 0.5em minus
  0.4em\relax IEEE, 2017, pp. 1639--1645.

\bibitem{lathuiliere2018deep}
S.~Lathuili{\`e}re, B.~Mass{\'e}, P.~Mesejo, and R.~Horaud, ``Deep
  reinforcement learning for audio-visual gaze control,'' in \emph{2018
  IEEE/RSJ International Conference on Intelligent Robots and Systems
  (IROS)}.\hskip 1em plus 0.5em minus 0.4em\relax IEEE, 2018, pp. 1555--1562.

\bibitem{gaolearning}
Y.~Gao, F.~Yang, M.~Frisk, D.~Hernandez, C.~Peters, and G.~Castellano,
  ``Learning socially appropriate robot approaching behavior toward groups
  using deep reinforcement learning,'' in \emph{International Conference on
  Robot {\&} Human Interactive Communication (RO-MAN)}.\hskip 1em plus 0.5em
  minus 0.4em\relax New Delhi: IEEE, 2019.

\bibitem{Meng2020}
L.~Meng, D.~Lin, A.~Francey, R.~Gorbet, P.~Beesley, and D.~Kuli{\'{c}},
  ``{Learning to Engage with Interactive Systems},'' \emph{ACM Transactions on
  Human-Robot Interaction}, vol.~10, no.~1, 2020.

\bibitem{Azar2017MinimaxRB}
M.~G. Azar, I.~Osband, and R.~Munos, ``Minimax regret bounds for reinforcement
  learning,'' in \emph{ICML}, 2017.

\bibitem{ziparo2008petri}
V.~A. Ziparo, L.~Iocchi, D.~Nardi, P.~F. Palamara, and H.~Costelha, ``Petri net
  plans: a formal model for representation and execution of multi-robot
  plans,'' in \emph{Proceedings of the 7th international joint conference on
  Autonomous agents and multiagent systems-Volume 1}.\hskip 1em plus 0.5em
  minus 0.4em\relax International Foundation for Autonomous Agents and
  Multiagent Systems, 2008, pp. 79--86.

\bibitem{Iocchi2016Pratical}
\BIBentryALTinterwordspacing
L.~Iocchi, L.~Jeanpierre, M.~Lazaro, and A.-I. Mouaddib, ``A practical
  framework for robust decision-theoretic planning and execution for service
  robots,'' \emph{Proceedings of the International Conference on Automated
  Planning and Scheduling}, vol.~26, no.~1, pp. 486--494, Mar. 2016. [Online].
  Available: \url{https://ojs.aaai.org/index.php/ICAPS/article/view/13790}
\BIBentrySTDinterwordspacing

\bibitem{Bubeck2012}
S.~Bubeck and N.~Cesa-Bianchi, ``{Regret analysis of stochastic and
  nonstochastic multi-armed bandit problems},'' \emph{Foundations and Trends in
  Machine Learning}, vol.~5, no.~1, pp. 1--122, 2012.

\bibitem{Jaksch2008NearoptimalRB}
T.~Jaksch, R.~Ortner, and P.~Auer, ``Near-optimal regret bounds for
  reinforcement learning,'' in \emph{J. Mach. Learn. Res.}, 2008.

\bibitem{peters2005model}
C.~Peters, C.~Pelachaud, E.~Bevacqua, M.~Mancini, and I.~Poggi, ``A model of
  attention and interest using gaze behavior,'' in \emph{International Workshop
  on Intelligent Virtual Agents}.\hskip 1em plus 0.5em minus 0.4em\relax
  Springer, 2005, pp. 229--240.

\end{thebibliography}
\bibliographystyle{IEEEtran}

\end{document}